\documentclass{article}

\usepackage[preprint]{neurips_2023}

\usepackage[utf8]{inputenc} 
\usepackage[T1]{fontenc}    
\usepackage{hyperref}       
\usepackage{url}            
\usepackage{booktabs}       
\usepackage{amsfonts}       
\usepackage{nicefrac}       
\usepackage{microtype}      
\usepackage{xcolor}         
\usepackage{graphicx}

\title{On the Amplification of Linguistic Bias through Unintentional Self-reinforcement Learning by Generative Language Models --- A Perspective}

\author{%
  Minhyeok Lee \\
  School of Electrical and Electronics Engineering\\
  Chung-Ang University \\
  Seoul 06974, Republic of Korea \\
  \texttt{mlee@cau.ac.kr} \\
}

\begin{document}

\maketitle

\begin{abstract}
Generative Language Models (GLMs) have the potential to significantly shape our linguistic landscape due to their expansive use in various digital applications. However, this widespread adoption might inadvertently trigger a self-reinforcement learning cycle that can amplify existing linguistic biases. This paper explores the possibility of such a phenomenon, where the initial biases in GLMs, reflected in their generated text, can feed into the learning material of subsequent models, thereby reinforcing and amplifying these biases. Moreover, the paper highlights how the pervasive nature of GLMs might influence the linguistic and cognitive development of future generations, as they may unconsciously learn and reproduce these biases. The implications of this potential self-reinforcement cycle extend beyond the models themselves, impacting human language and discourse. The advantages and disadvantages of this bias amplification are weighed, considering educational benefits and ease of future GLM learning against threats to linguistic diversity and dependence on initial GLMs. This paper underscores the need for rigorous research to understand and address these issues. It advocates for improved model transparency, bias-aware training techniques, development of methods to distinguish between human and GLM-generated text, and robust measures for fairness and bias evaluation in GLMs. The aim is to ensure the effective, safe, and equitable use of these powerful technologies, while preserving the richness and diversity of human language.
\end{abstract}

\section{Introduction}

As we navigate the digitized age, we increasingly witness the penetration of generative language models (GLMs) into a multitude of sectors \citep{openai2023gpt4,radford2019language,radford2018improving,brown2020language}. These advanced artificial intelligence tools are being harnessed to craft a broad array of written content; from concocting engaging novels, developing intriguing scripts, to composing relevant news pieces from mere topical cues. This expanding footprint of generative language models in the realm of writing presents an intriguing shift in how we create and consume text, marking a potential turning point in the annals of written communication.

In addition to the impact of GLMs in written content creation, it is important to acknowledge the role of ChatGPT in the widespread adoption of these tools. ChatGPT \citep{openai2023gpt4}, a powerful language model developed by OpenAI, has played a significant role in expanding the reach and accessibility of AI-generated text. With its ability to understand and respond to human prompts, ChatGPT has become a potential tool for various applications, including customer service, virtual assistants, and interactive conversations.

The proliferation of written pieces crafted, or at least aided, by these generative models is inevitable given their growing acceptance and widespread adoption. The convenience and efficiency offered by these tools have seen them embraced in numerous applications, bringing about a transformation in content generation processes. Yet, it is not without concern. In particular, the possibility that these generative language models might unintentionally amplify biases present in the data they were trained on, raises critical questions.

The potential risk of self-reinforcing bias in learning from these models poses a significant challenge. As more content is generated by these models, and in turn, used to train future iterations, the opportunity for biases to be reinforced and perpetuated increases. This phenomenon could lead to a spiral of bias amplification, undermining the objectivity and fairness of the content produced, a factor that necessitates thorough scrutiny.

Our exploration here is motivated by the need to better understand these implications, with a focus on unintended self-reinforcement learning. We posit that an unchecked adoption and uncritical consumption of content generated by these models could inadvertently facilitate an echo chamber of biases. In the following sections, we shall explore this issue, discussing potential ramifications and suggesting possible countermeasures to mitigate this unintended consequence.

\section{Linguistic Bias in Generative Language Models}

Before exploring the potential consequences of bias propagation in GLMs, it is essential to define the type of bias that is the focus of our discussion in this paper. Bias, in the context of machine learning and more specifically, GLMs, can manifest in numerous ways. However, in this work, we predominantly concentrate on the linguistic biases in vocabulary selection, tone, and the flow of writing.

In the field of machine learning, bias generally refers to a pattern of errors that the model tends to make based on the training data it was provided \citep{mehrabi2021survey}. While this notion of bias holds in our context as well, we take a step further to highlight the more subtle, less noticeable, and yet, profoundly impactful dimensions of bias that emerge in the outputs of GLMs.

Before exploring the potential consequences of bias propagation in GLMs, it is essential to define the type of bias that is the focus of our discussion in this paper. Bias, in the context of machine learning and more specifically, GLMs, can manifest in numerous ways. However, in this work, we predominantly concentrate on the linguistic biases in vocabulary selection, tone, and the flow of writing.

While some biases, such as non-linguistic biases, might seem more egregious or apparent, we argue that they may not pose as substantial a problem in the context of GLMs. This is due to the fact that users of these models often act as primary filters for explicit biases. Non-linguistic biases typically manifest as overt preferences towards a particular gender, race, or minority, or they may involve the production of inaccurate information or hallucinations \citep{math11102320}. When GLMs generate content exhibiting such biases, human users can usually identify and discard these skewed outputs, preventing this biased information from proliferating in the public domain.

On the contrary, the more pervasive yet subtle problem lies in the linguistic bias inherent in the operation of GLMs. These models function by predicting the next word based on the context provided by preceding tokens or words. In making this prediction, the model relies heavily on probability distributions that have been learned from the training data. Given a particular token or word, the model is programmed to select the next word from those with the highest associated probabilities. This process leads to a pattern where, given a similar context or initial token, the model tends to produce similar outputs over repeated instances.

What this implies is that, over time, GLMs can end up generating text that displays a consistent bias in terms of word selection, tone, and flow of writing, as these are the patterns that the model has learned from the training data. The issue with this form of linguistic bias is that it is far more difficult for users to detect, let alone filter. Unlike non-linguistic biases that are often explicit and recognizably deviant, linguistic biases are nuanced, blending seamlessly into the fabric of the text, making it hard for users to even realize their presence. 

Vocabulary selection bias refers to the model's preference for certain words, phrases, or terminologies over others. This bias could be a result of the frequency of word usage in the training data, or it could reflect the prominence of certain cultural, social, or professional contexts in the data pool. Tonal bias involves the tendency of a model to reflect a particular style, emotion, or sentiment in its output. A GLM trained predominantly on formal, academic literature may naturally adopt a more formal tone, even when generating text for informal or casual contexts. Conversely, a GLM trained on a data pool laden with slang or colloquial language might incorporate such elements even in contexts requiring formal expression. The flow of writing is another aspect where bias can emerge. This could involve the structure of sentences, the sequence of ideas, the use of transition words, and the overall cohesiveness and coherence of the generated text. Models may develop a default writing flow based on the predominant patterns in their training data, which can subsequently influence the way ideas are organized and communicated in the generated text.

These types of bias, while not overtly harmful or discriminatory, can still have substantial influence over how language is perceived and utilized. They are subtle, often going unnoticed, and yet, their impact on language development, communication, and the representation of knowledge can be profound.

It is important to note that such biases are not a product of intentional design decisions. Rather, they emerge as a consequence of the training data's characteristics and the model's learning mechanisms. The biases in vocabulary, tone, and writing flow are reflections of the patterns the models have learned from their training data, and they are perpetuated and reinforced when these models generate text, which then serves as training data for subsequent models.

\section{Possibility of Unintended Self-reinforcement Learning of GLMs} \label{sec:training}

GLMs, with their adaptive learning capabilities and extensive application, can shape the textual landscape of the internet in significant ways. However, the effectiveness of these models hinges largely on the dataset they learn from. The more diverse and comprehensive the dataset, the better equipped the model is in delivering accurate and contextually correct content \citep{tirumala2022memorization}.

The issue arises when GLMs, as part of their learning process, use internet-crawled data to fine-tune their capabilities. As GLMs gain popularity and their outputs multiply across the digital space, these AI-generated texts inevitably become a part of the data pool from which future GLMs learn. While this seems like a benign cycle, it presents a unique challenge, the potential for self-reinforcement of biases.

GLMs, similar to any machine learning models, are susceptible to biases present in their training data. These biases could be related to language use, tone, gender, race, knowledge accuracy, or any other facet of human discourse and knowledge representation. If an initial GLM has a certain bias, it is likely that the text it generates will also exhibit the same bias. As this biased content contributes to the learning material for subsequent models, the bias can unintentionally be reinforced and amplified, leading to a self-perpetuating cycle.

The self-reinforcing nature of learning in GLMs can thus create an echo chamber of sorts, where biases are amplified rather than mitigated with each iteration of learning. Given the difficulty in distinguishing human-written text from GLM-generated text, these biases can easily propagate unimpeded.

This inadvertent self-reinforcement is particularly potent when it comes to linguistic biases. As previously discussed, such biases, rooted in vocabulary selection, tonal preferences, and writing flow, are considerably subtle and often blend into the generated text without causing overt disruptions. Consequently, these biases are not easy for users or automated filters to detect and correct, which allows them to permeate the generated text and, in time, the overall linguistic landscape shaped by GLMs.

Furthermore, given the expansive and ubiquitous use of GLMs, such as chatbots, automated content creators, and translators, these biases can diffuse across a vast array of contexts. When GLM outputs, with their embedded linguistic biases, become a part of the learning pool for subsequent models, we encounter the risk of these biases becoming deeply entrenched and normalised. This could, in effect, lead to the unintentional shaping of language use and expression in the digital domain, to mirror the biases of the models, rather than a balanced representation of human language and thought.

This cycle of learning, generation, and re-learning effectively serves as an unintentional feedback loop, where the biases in the initial output are fed back into the learning system, only to emerge stronger in the subsequent output. Without appropriate measures in place to identify and correct these biases, they can continue to propagate through this loop, growing more pronounced with each iteration.

\begin{figure}[t]
    \centering
    \includegraphics[width=\textwidth]{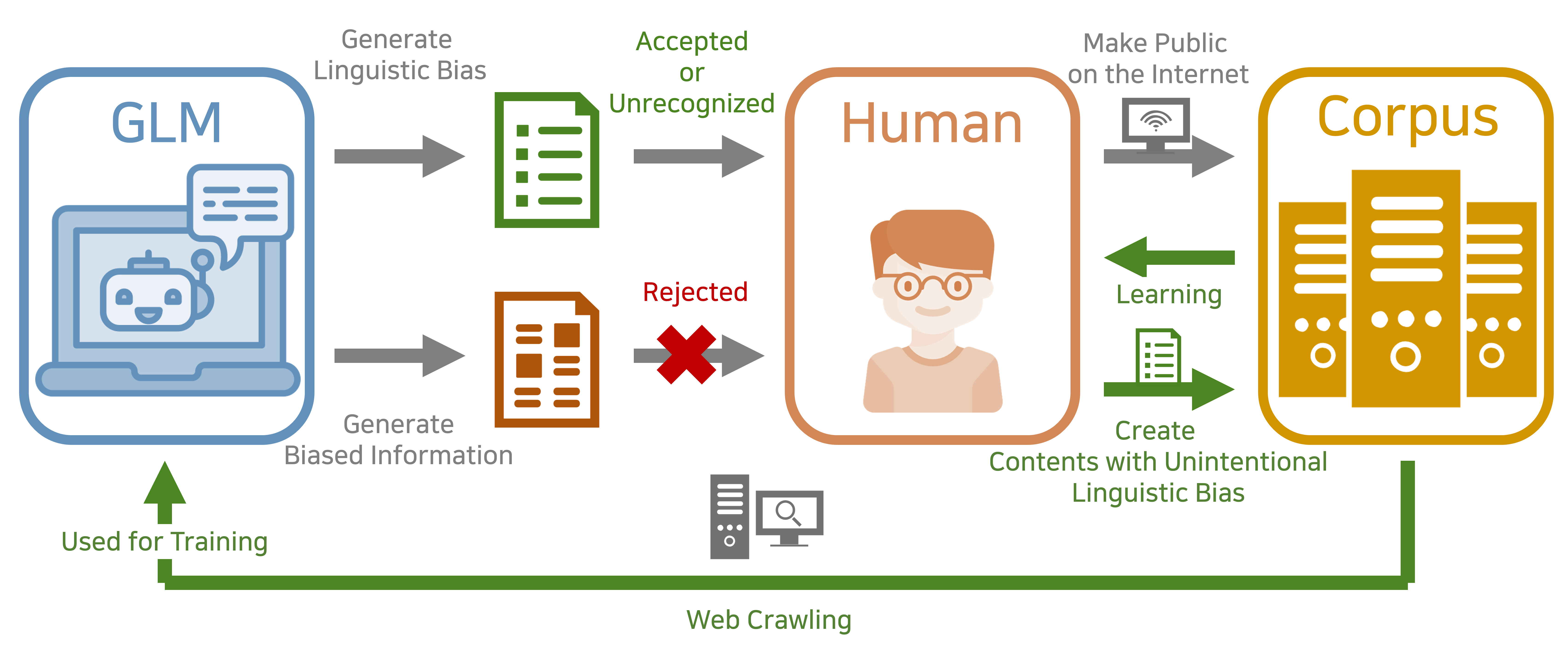}
    \caption{\textbf{Process of Unintentional Self-reinforcement Learning of Linguistic Bias.} Green lines represent the process of self-reinforcement learning of linguistic bias}
    \label{fig:learning}
\end{figure}

\section{Possibility of Unintended Bias Learning by Humans} \label{sec:learning}

As GLMs increasingly become intertwined with our digital interactions, it is crucial to contemplate not just the self-reinforcing bias within the models themselves, but also the inadvertent bias that may be learned by humans. This is particularly relevant in the context of education, where the internet has assumed a significant role. It is thus necessary to scrutinize the subtle ways in which GLM-generated content can impact the linguistic and cognitive development of future generations.

False information or blatantly biased writing is generally identifiable and hence, can be avoided or corrected in educational contexts. However, the challenge lies with the subtler, more pervasive forms of bias, those which are not glaringly wrong but are nonetheless, influential. These could be biases in vocabulary selection, tone, or the flow of writing that emerge from the GLMs. Given the widespread use of the internet in advanced education, children are likely to be exposed to and subsequently, unconsciously absorb these biases, impacting their linguistic growth and their interpretation of the world.

This inadvertent learning of bias has implications for the evolution of language and communication. The vocabulary choices, tonal preferences, and writing styles of early GLMs could seep into the lexicon of future generations, potentially shaping and constraining how they express ideas and emotions. This phenomenon, while not harmful in a traditional sense, still represents a significant influence on the natural development and diversity of language.

Furthermore, once these GLM-induced biases are learned and reproduced by humans, they become embedded in human-generated text. Consequently, even if future GLMs are trained primarily on human-written text, they are still likely to encounter and learn from the biases of earlier GLMs. This presents a paradoxical situation where attempting to reduce the influence of GLMs by focusing on human-written content might inadvertently sustain the very biases we seek to eliminate.

Thus, the widespread adoption of GLMs may catalyze a loop of bias reinforcement and amplification that extends beyond the models themselves, permeating human language and communication. This underscores the importance of addressing biases in GLMs, not just to improve the models, but also to protect the diversity and integrity of human language and discourse. Figure~\ref{fig:learning} illustrates the process of unintentional self-reinforcement learning of linguistic bias during human learning and GLM training.

\section{Pros and Cons of Linguistic Bias Amplification in Widespread Adoption of GLMs}

\subsection{Advantages}

\textbf{Educational Benefits} The first major advantage is tied to the domain of child education. The statistical nature of GLMs allows them to generate logical and rational text from a linguistic point of view. This outcome is a byproduct of their training, where they learn from vast corpora to generate text that closely mirrors high-quality human language \citep{lee2023mathematical}. As such, the output of GLMs can be beneficial in an educational context. Children learning from GLM-generated text can potentially acquire enhanced writing skills due to exposure to rational and well-constructed language. The inherent linguistic bias of the GLMs, in this case, might result in the propagation of refined language skills.

\textbf{Facilitation of Future GLM Learning} The second advantage lies in the ease of learning for future GLMs. Under the premise that the linguistic bias inherent in the text generated by current GLMs is diffused widely and utilized in subsequent GLM training, this can essentially expedite the learning process of future GLMs. This can be attributed to the principle of knowledge distillation in deep learning \citep{hinton2015distilling}. Knowledge distillation refers to the process wherein one deep learning model is trained using the outputs of another, allowing the model to assimilate key data and knowledge in a relatively shorter span of time \citep{gou2021knowledge,cho2019efficacy}. In this context, despite the potential linguistic bias, the core information embedded in the training data is preserved in the generated text. Consequently, this not only sustains the essence of the original information but also expedites the training of subsequent GLMs, rendering the learning process more efficient and swift. Hence, the linguistic bias, while it may be perceived as a challenge, might conversely serve as an enabler in streamlining the learning of future GLMs.

\subsection{Disadvantages}

\textbf{Reduction of Linguistic Diversity} Conversely, there are potential drawbacks to the widespread adoption of GLMs, one of which concerns the preservation of human linguistic diversity. Language is a fundamental aspect of human culture and identity, and it is inherently diverse and evolving. If GLM-produced text, with its inherent linguistic bias, becomes a predominant source of language learning for humans, there is a risk of these biases being unconsciously absorbed and propagated by human language users. This could lead to a scenario where human language begins to mirror the language of GLMs, leading to a reduction in the diversity and richness of human language. Moreover, considering the Sapir-Whorf Hypothesis which postulates that the structure of a language affects its speakers' world view \citep{o2015sapir,kay1984sapir}, a reduction in linguistic diversity might also impact the diversity of human thought.

\textbf{Dependency on Initial GLMs} The second disadvantage pertains to the dependence on initial GLMs. As we have discussed, the biases inherent in the output of the currently developed GLMs like ChatGPT could potentially propagate through subsequent GLMs and human language, creating a self-reinforcing cycle of bias. This introduces a dependency on the initial GLMs, a situation which might be disadvantageous due to several reasons. First, it could limit the flexibility and adaptability of future GLMs and human language, as they are constrained by the linguistic biases of the initial GLMs. Second, this early GLM-dependency scenario makes it difficult to predict the future impacts of GLMs without sufficient research and efforts to mitigate their current linguistic biases. Until these biases are adequately understood and addressed, the self-reinforcing cycle might continue, further exacerbating these problems.

\section{Challenges and Future Directions}

The exploration of the benefits and drawbacks of linguistic biases present in GLMs highlights the complexity of their widespread use. While GLMs promise numerous advantages, their latent ability to unintentionally reinforce and amplify inherent biases calls for rigorous scrutiny. To ensure the effective, safe, and fair application of these technologies, a comprehensive understanding of these issues, coupled with dedicated efforts to address them, is of paramount importance.

\subsection{Challenges}

The primary challenge lies in the dependency on initial GLMs. The current generation of GLMs, such as ChatGPT, inadvertently influence future models and human language through their intrinsic linguistic biases. This interdependency poses several difficulties. On the one hand, it restricts the dynamism of subsequent GLMs and human language, as both are perpetually influenced by the biases of the initial models. On the other hand, it obscures the foreseeability of the impacts of GLMs, as the extent and depth of the biases in current GLMs have not yet been fully explored or understood.

The propagation of linguistic bias is a subtle and complex process, which makes it difficult to detect and quantify. Unlike explicit misinformation or overt bias, linguistic bias often manifests in nuanced ways, such as a preference for certain vocabulary or syntax, or a skew towards a particular tone or style. These biases, while they might appear trivial, can have significant consequences when they are unknowingly adopted and perpetuated by both humans and machines.

\subsection{Future Directions}

Given these challenges, future work should be directed towards a more profound understanding of the linguistic biases inherent in current GLMs. Thorough research should be conducted to identify, quantify, and comprehend the extent of these biases. This knowledge would not only shed light on the possible impacts of GLMs on human language and thought, but also guide the development of strategies to mitigate these biases.

Efforts should also be made to improve the transparency and interpretability of GLMs. The underlying mechanisms that give rise to linguistic bias should be dissected, and the reasons why certain biases are more likely to be amplified than others should be investigated. This could be accomplished through techniques such as feature importance analysis or visualization of hidden layer activations.

In addition, strategies should be developed to prevent the reinforcement and amplification of linguistic bias. These could include techniques for bias-aware model training, such as bias regularization or bias-attentive data augmentation. Alternatively, post-processing methods could be employed to adjust the output of GLMs to reduce bias.

Moreover, there should be ongoing work to develop techniques to distinguish between human-written text and GLM-generated text, in order to prevent the unintentional ingestion of GLM-generated text into the training data of future GLMs. The development of such techniques could help to curb the self-reinforcement of linguistic bias.

It is essential to develop robust and rigorous measures to evaluate the fairness and bias of GLMs. These measures should not only consider the output of the models but also the processes that generate these outputs. By holding GLMs to high standards of fairness and bias, we can ensure that they contribute positively to our linguistic landscape, rather than distort it with amplified bias.

\section{Conclusion}

The widespread adoption of GLMs holds profound implications for our digital interactions and the evolution of human language. As GLMs like ChatGPT are increasingly utilized across various applications, their influence on shaping the digital linguistic landscape is undeniable. However, this influence brings with it the latent potential for the unintentional reinforcement and amplification of inherent linguistic biases. These biases, rooted in the initial training data of the GLMs, might not only propagate through subsequent models but also infiltrate human language and thought, leading to a self-reinforcing cycle of bias amplification.

While the utility of GLMs in areas such as education and the facilitation of future GLM learning is commendable, the inadvertent propagation of linguistic biases poses significant challenges. The consequences of these biases are multifaceted, ranging from the reduction of linguistic diversity to the creation of an unintended dependency on initial GLMs. Recognizing these issues is crucial to foreseeing and preparing for the potential impacts of widespread GLM usage.

Addressing this complex situation demands a robust research approach aimed at understanding and mitigating the biases in current GLMs. Future research should be directed towards uncovering the nature and extent of these biases, improving model transparency, and devising methods to prevent bias amplification. It is crucial to develop techniques to distinguish between human-written and GLM-generated text and establish robust measures for evaluating GLM fairness and bias.

The journey towards bias-free GLMs is undoubtedly challenging but essential for the future of digital interactions and the preservation of the richness and diversity of human language. By investing in rigorous research and taking conscientious strides towards mitigating these biases, we can unlock the true potential of GLMs in a manner that is not only technologically beneficial but also ethically sound. In so doing, we can ensure that the widespread adoption of GLMs bolsters the diversity of human language and thought, rather than unintentionally constraining them. Thus, the path towards this vision, while demanding, promises to be a crucial undertaking for the future of artificial intelligence and its coexistence with humanity.

\begin{ack}
We would like to bring to your attention that the refinement of this paper, achieved through the utilization of GLMs, may have unintentionally introduced linguistic bias into the main text.
\end{ack}

\bibliographystyle{unsrtnat}
\bibliography{ref.bib}

\end{document}